\documentclass[a4paper, 10 pt, conference]{ieeeconf} 

\IEEEoverridecommandlockouts                              % This command is only needed if 
\usepackage{amsmath} % assumes amsmath package installed
\usepackage{amssymb}  % assumes amsmath package installed
\usepackage{graphicx}
\usepackage{subcaption}
\usepackage{booktabs}
\usepackage{multirow}
\usepackage{glossaries}
\usepackage{hyperref}
\usepackage{xurl}
\hypersetup{breaklinks=true}
\usepackage{xcolor}
\usepackage{soul}
\usepackage[english]{babel}
\usepackage[autostyle, english=british]{csquotes}
\usepackage{censor}
\StopCensoring

\usepackage[natbib=true, style=ieee, autocite=plain, maxcitenames=2, isbn=false, backend=biber]{biblatex}
\DeclareSourcemap{
  \maps[datatype=bibtex]{
    \map{
      \step[fieldsource=doi, final]
      \step[fieldset=url, null]
      \step[fieldset=urldate, null]
    }
    \map{
      \step[fieldsource=howpublished, final]
      \step[fieldset=type, null]
    }
  }
}
\AtEveryBibitem{\clearfield{day}}

\newacronym{usv}{USV}{Unmanned Surface Vehicle}
\newacronym{sci}{SCI}{Smallest Credible Interval}
\newacronym{neu}{NEU}{North, East, Up}
\newacronym{slam}{SLAM}{Simultaneous Localisation and Mapping}
\newacronym{ipp}{IPP}{Informative Path Planning}
\newacronym{ros}{ROS}{Robot Operating System}
\newacronym{cfd}{CFD}{Computational Fluid Dynamics}
\newacronym{kld}{KLD}{Kullback-Leibler Divergence}
\newacronym{cimb}{CIMB}{collectively independent multivariate Bernoulli}
\newacronym{anova}{ANOVA}{analysis of variance}

\title{\LARGE \bf
Active Tracking of Marine Pollution Sources: An Uncertainty-Aware Categorical Bayesian Framework for Unmanned Surface Vehicles*
}

\xblackout{
\author{Song Ma$^{1}$, Yanchao Wang$^{1}$, Yewei Huang$^{2}$, Richard Bucknall$^{1}$, and Yuanchang Liu$^{1}$% <-this % stops a space
\thanks{*This work was supported by the Dean's Prize of UCL Engineering; the China Scholarship Council; Engineering and Physical Sciences Research Council (EPSRC) [grant number EP/Y000862/1]; the Royal Society Kan Tong Po Fellowship (KTP/R1/251117).}% <-this % stops a space
\thanks{$^{1}$Department of Mechanical Engineering, University College London, Torrington Place, London, WC1E 7JE, UK.
        {\tt\small \{song.ma.18, yanchao.wang.20, r.bucknall, yuanchang.liu\}@ucl.ac.uk}}%
\thanks{$^{2}$Dartmouth College, Hanover, NH 03755, USA.
        }
}
}

\AtBeginDocument{
    \hypersetup{
        pdfsubject={Marine Pollution Source Tracking},
        pdfkeywords={Decision Making, Autonomous Systems, marine robotics, Informative Path Planning},
        pdfauthor={Ma, Song},    
        pdftitle={Active Tracking of Marine Pollution Sources:An Uncertainty-Aware Categorical Bayesian Framework for Unmanned Surface Vehicles},
    }
}

\begin{document}

\maketitle
% \IEEEpeerreviewmaketitle

\thispagestyle{empty}
\pagestyle{empty}

%%%%%%%%%%%%%%%%%%%%%%%%%%%%%%%%%%%%%%%%%%%%%%%%%%%%%%%%%%%%%%%%%%%%%%%%%%%%%%%%
\begin{abstract}
This paper presents an uncertainty-aware framework for the active tracking of marine pollution sources using Unmanned Surface Vehicles (USVs).
The proposed framework employs an Informative Path Planning (IPP) strategy driven by Bayesian inference, modelling the belief of source location as a categorical distribution. This work presents a high-fidelity simulation pipeline, coupling Computational Fluid Dynamics (CFD) for realistic pollutant dispersion with Gazebo-based hydrodynamics and ArduPilot for USV control. Furthermore, this paper introduces the Smallest Credible Interval (SCI) as a metric to quantify estimation uncertainty and to serve as an autonomous termination criterion. Extensive simulations across diverse wave conditions and source locations demonstrate that the proposed framework achieves a 95.8\% success rate, significantly outperforming baseline methods in both localisation accuracy and environmental adaptability. This framework provides a scalable and ROS-compatible foundation for fully autonomous environmental monitoring and rapid incident response.
\end{abstract}

%%%%%%%%%%%%%%%%%%%%%%%%%%%%%%%%%%%%%%%%%%%%%%%%%%%%

\section{Introduction}

% \subsection{Background}
Pollution discharged into the marine environment causes severe consequences to ecosystems \autocite{
    % johnston_review_2015, 
    tornero_chemical_2016} and human health \autocite{landrigan_human_2020}. The marine pollution problem arises from various sources. According to \textcite{biswas_coastal_2021}, coastal waters are endangered by sewerage discharge, agricultural, and industrial waste. Even in the open ocean, water bodies face hazards like oil spills and chemical leaks from various maritime activities \autocite{solberg_remote_2012}. To minimise the influence of marine pollution, reliable monitoring and rapid responses play a crucial role.

In order to better monitor marine pollution and restrict its influence, the use of robotics systems has attracted much attention, which reduces the risks that humans are exposed to \autocite{brantner_controlling_2021}. In particular, the use of robotic systems in environmental monitoring has been widely investigated \autocite{trincavelli_towards_2008}, and it demonstrates significantly higher efficiency compared with conventional solutions \autocite{fahimi_autonomous_2009}. A crucial sub-problem within robotic marine monitoring is the task of localising the pollution source, often referred to as source tracking \autocite{jing_recent_2021}.

% \glspl{usv} has attracted much attention. Deploying \glspl{usv} reduces the risks that humans are exposed to \cite{brantner_controlling_2021}. Furthermore, it demonstrates the potential for significantly higher efficiency when utilising \glspl{usv} with autonomous decision-making systems \cite{fahimi_autonomous_2009}. A major current focus in pollution monitoring using autonomous systems is how to perform active pollution source tracking.
% \begin{figure}[htbp]
% \centerline{\includegraphics[width=0.9\linewidth]{figs/concept.pdf}}
% \caption{Survey USV cruising in high-fidelity simulated environment, searching for pollution source.}
% \label{fig:survey_usv}
% \end{figure}

% \subsection{Related Research}

Early approaches to pollution source tracking focused on establishing estimation models of the source locations. For example, \textcite{pang_chemical_2006} used Bayesian inference to generate a probability map of the source location in a marine environment, while \textcite{hutchinson_source_2019} applied a similar principle to airborne source localisation using a particle filter. A key limitation of these methods is their reliance on external control inputs, such as a predefined search pattern (e.g., a lawnmower pattern), to collect data. Consequently, these methods are not fully autonomous and can be considered as a variant of deploying a fixed sensor network \autocite{liu_bayesian_2024}.

To overcome this reliance on predefined paths and achieve full autonomy, research has shifted towards active control strategies, often referred to as active information gathering or \gls{ipp} \autocite{chen_pareto_2019}. Several researchers have proposed to incorporate the \gls{ipp} to substitute the external control in the fields of localisation \autocite{fox_active_1998}, \gls{slam} \autocite{placed_survey_2023}, and 3D reconstruction \autocite{zhu_online_2021}. Infotaxis \autocite{vergassola_infotaxis_2007} firstly introduced the concept of using information gain to guide source tracking, inspired by the chemotaxis of certain organisms. More recent works \autocite{ojeda_information-driven_2021, rhodes_autonomous_2022} further extend the studies of infotaxis to gas source localisation and gas source term estimation. With research works focusing on various search strategies, few have investigated the effect of the probabilistic models used in the algorithms.

On the application aspect, although literature like \textcite{bayat_optimal_2016} put forward active source tracking for marine pollution, compared to the field of indoor gas source localisation, more work is needed to tackle the source tracking for marine pollution. The relatively fewer works can be a result of the difficulties in conducting simulated and experimental validations for marine environments. Existing high-fidelity underwater robotic simulators \autocite{manhaes_uuv_2016,prats_open_2012} and surface robotic simulators \autocite{bingham_toward_2019} are widely used in perception and manipulation tasks, while many \gls{ipp} algorithms still rely on abstract grid-based simulations due to a lack of a specialised pipeline.

\begin{figure*}[thbp]
\centerline{\includegraphics[width=\textwidth]{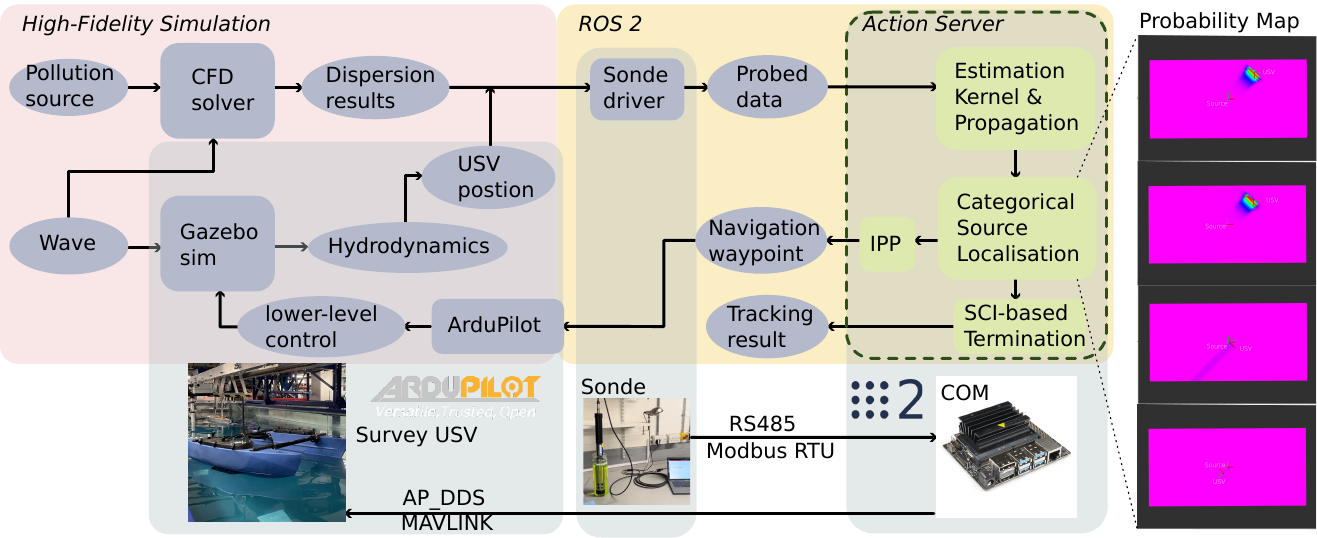}}
\caption{Overall workflow of the proposed framework and its associated hardware components.}
\label{fig:framework}
\end{figure*}

% \subsection{Major Contribution}
In the present paper, an uncertainty-aware active pollution source tracking framework using categorical Bayesian estimation is proposed. 
% The framework is \gls{ros}-compatible, allowing the proposed tracking system to be fully integrated into the simulated marine pollution data derived by \gls{cfd} results and the deployed robot was an environmental survey \gls{usv}.
We summarise the contributions of this work as follows:
\begin{itemize}
    \item The proposed framework enhances the underlying Bayesian inference by utilising a categorical distribution instead of the \gls{cimb} model. Additionally, it introduces the \gls{sci} as a metric to quantify real-time uncertainty and serve as an autonomous termination criterion.

    \item A specialised modular simulation pipeline for marine pollution monitoring using \glspl{usv} is established to validate the algorithm, bridging professional-standard software stacks, namely ROS 2, Gazebo, ArduPilot and OpenFOAM.

    \item Extensive validation in simulated marine environments demonstrated a 95.8\% overall success rate, with the proposed method significantly outperforming Poisson-Bessel Infotaxis and CIMB-based baselines in both accuracy and environmental robustness.
\end{itemize}

% \begin{itemize}
%     \item A high-fidelity simulation framework for marine pollution monitoring using \glspl{usv} is constructed. This \gls{ros}-compatible simulation featured a \gls{cfd}-based dispersion simulator and a physical simulator of the survey \gls{usv}.
%     \item An uncertainty-aware active pollution source tracking system is proposed. The system modelled the pollution source location as a categorical distribution and navigated the vehicle to progressively refine the estimate by maximising the expected information gain. The uncertainty level of the source location estimate was quantified using credible intervals.
%     \item The proposed framework is validated in various pollution dispersion scenarios within the high-fidelity simulated environment. The results demonstrate that the proposed method can reliably localise pollution sources with high accuracy and outperforms the existing baseline.
% \end{itemize}
Source code and configurations of the present paper can be found at this repository\footnote{
% \url{https://anonymous.4open.science/r/marine_source_tracking-3D74/}
\url{https://github.com/ucl-frl/marine_source_tracking/}
}.

\section{Marine pollution source tracking framework}
\label{s:framework}
The proposed framework consists of two modules: (1) a high-fidelity simulator for both marine pollution and the survey \gls{usv}; (2) a \gls{ros}-based source tracking system navigating the survey \gls{usv} through sending waypoints to the ArduPilot interface, as can be seen from Fig. \ref{fig:framework}. Such a simulation-driven framework plays a significant role in the development and validation processes of robotic software, as it allows for extensive testing without the need for physical hardware. This is particularly important in the context of marine pollution monitoring, where hardware testing can be challenging and expensive. The source tracking algorithm was wrapped as \gls{ros} packages, which enables an efficient switch from simulations to hardware implementations.
\subsection{Pollution dispersion and USV simulator}
\label{ss:simulator}
The simulation module includes two parts. First, a \gls{cfd} solver is used to compute and simulate the dispersion of the pollutant based on source input, $f\left( p \right)$, the initial concentration, $c_0$, and the wave velocity, $\mathbf{v}$. Second, a physical simulator of the survey \gls{usv} is used to process the control input and derive the hydrodynamic response of the \gls{usv}.

\begin{figure*}[htbp]
\centering
    \begin{subfigure}[t]{0.4\textwidth}
        \centering
        \includegraphics[width=\textwidth]{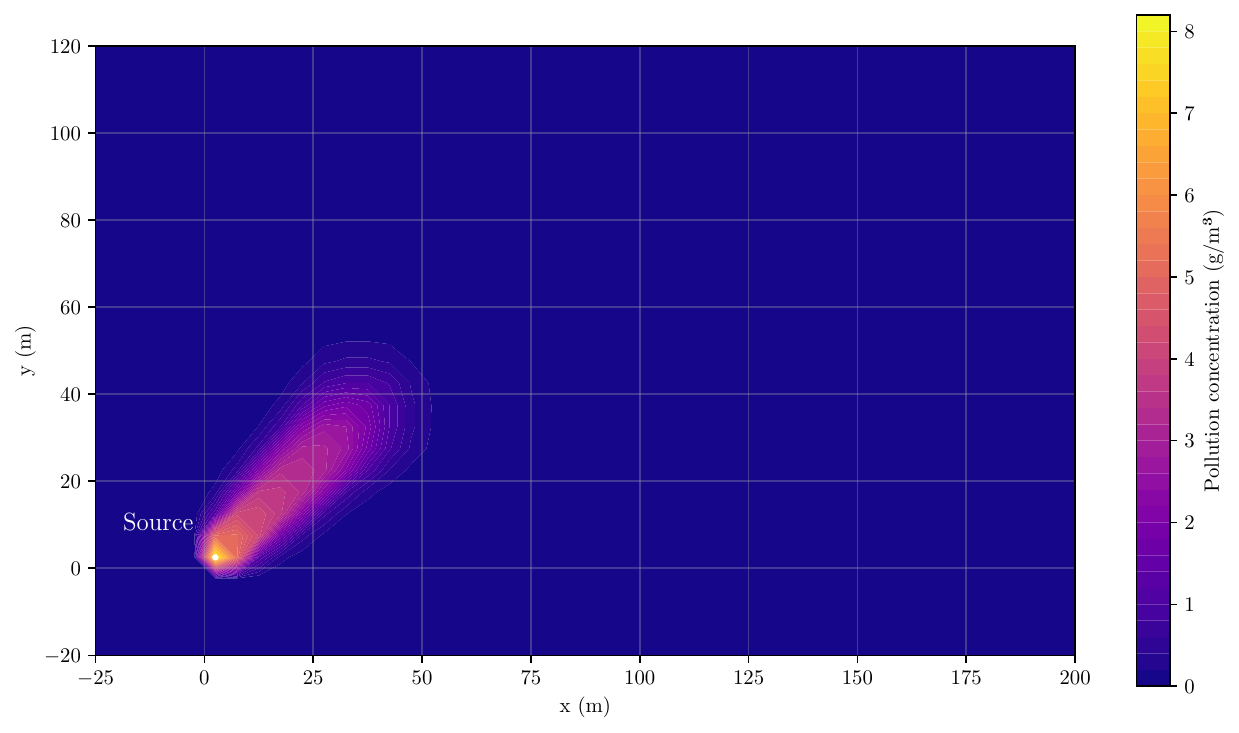}
        \caption{30 s}
        % \label{}
    \end{subfigure}
    \begin{subfigure}[t]{0.4\textwidth}
        \centering
        \includegraphics[width=\textwidth]{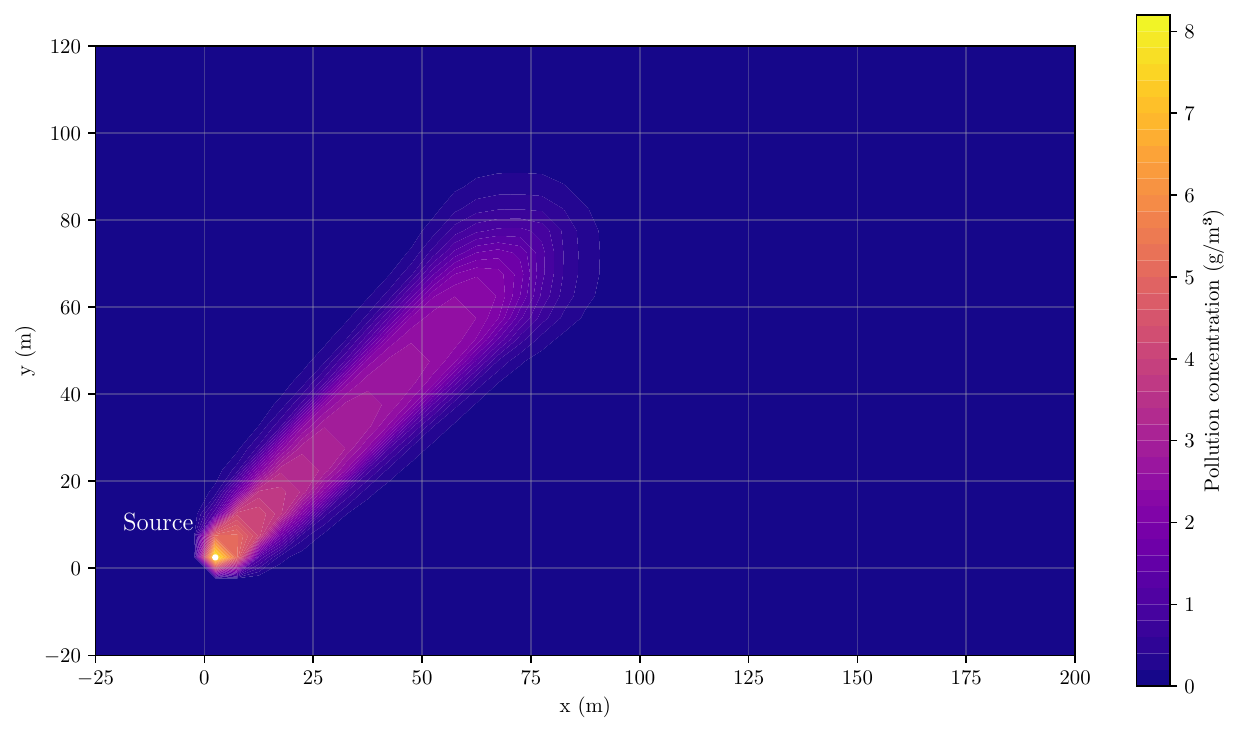}
        \caption{60 s}
        % \label{}
    \end{subfigure}
        \begin{subfigure}[t]{0.4\textwidth}
        \centering
        \includegraphics[width=\textwidth]{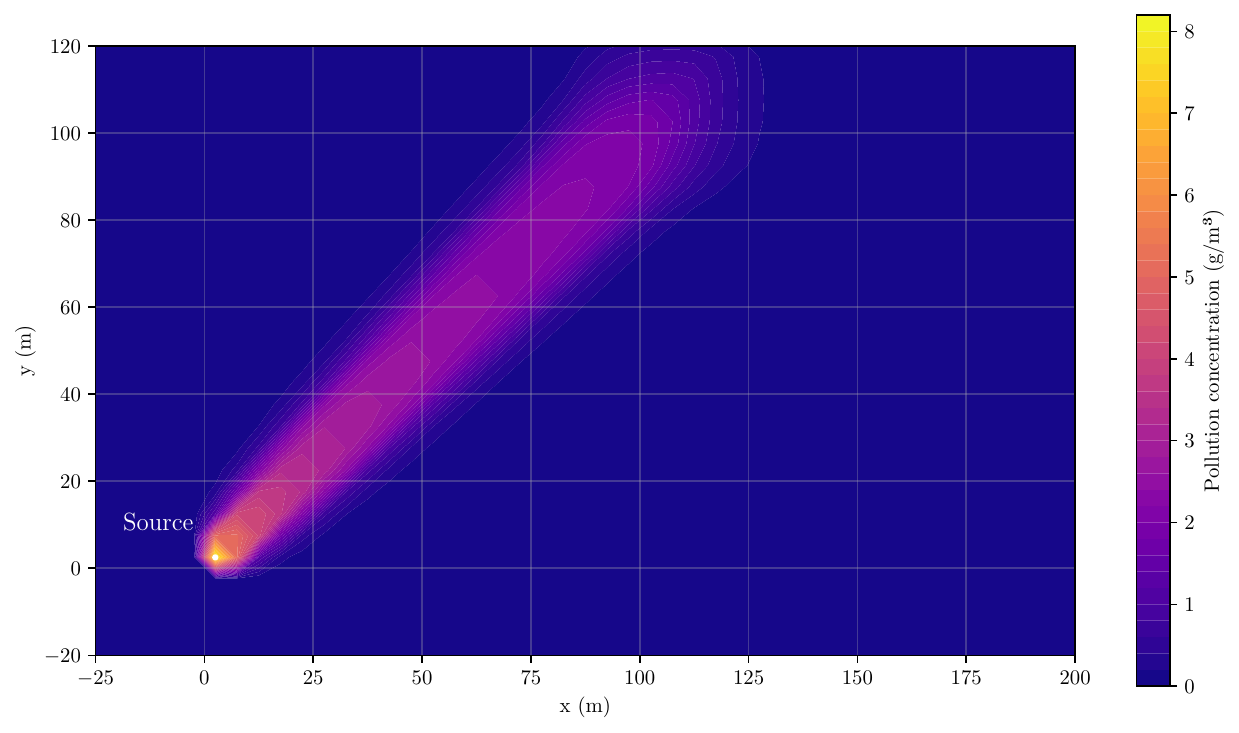}
        \caption{90 s}
        % \label{}
    \end{subfigure}
        \begin{subfigure}[t]{0.4\textwidth}
        \centering
        \includegraphics[width=\textwidth]{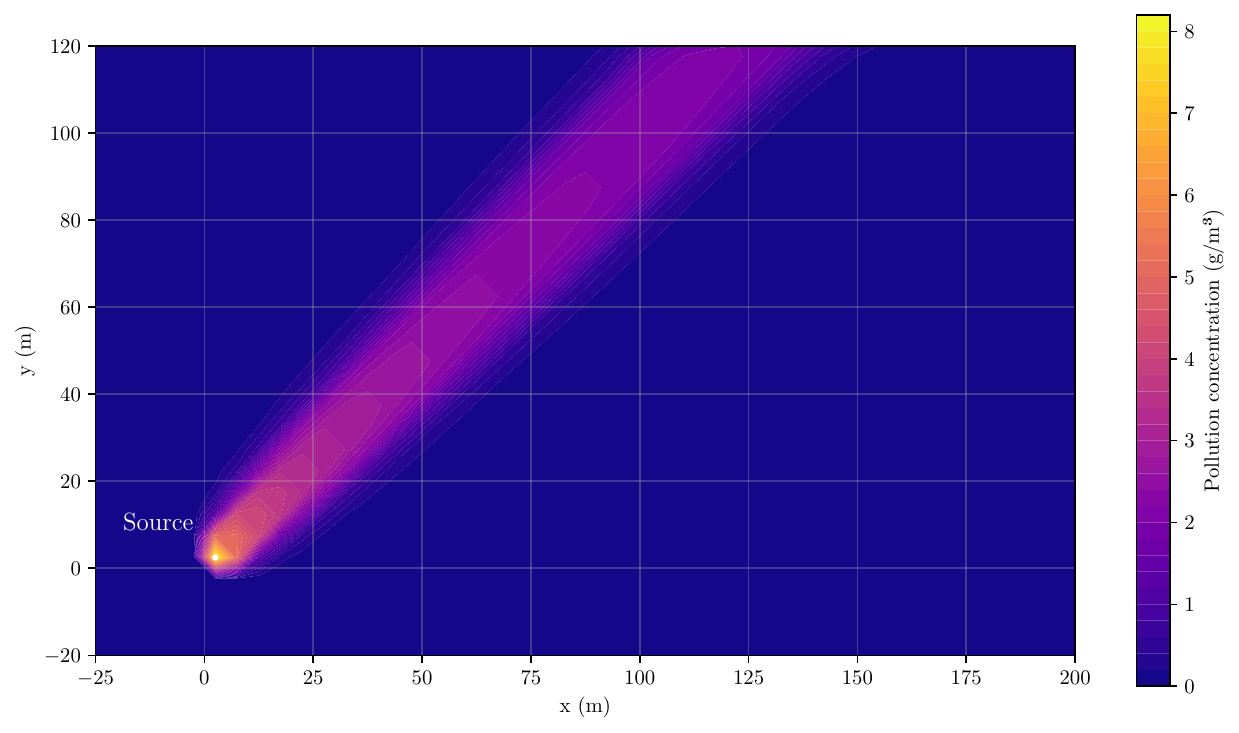}
        \caption{120 s}
        % \label{}
    \end{subfigure}
    \caption{The simulated time-varying pollution dispersion plume at different time steps. The wave velocity is $\mathbf{v} = \left[1.2247, 1.2247\right]^\intercal$ m/s.}
    \label{fig:dispersion}
\end{figure*}

The dispersion of the pollutant is modelled by the diffusion-advection equation as reported in \textcite{ma_adaptive_2025}. The diffusion-advection model has been used in numerous investigations \autocite{choy_advection-diffusion_2000, ulfah_advection-diffusion_2018} related to environmental monitoring. The diffusion-advection equation is given as:
\begin{equation}
    \dot{c}(p, t)-\lambda \nabla^2 c(p, t)+\nabla \cdot(\mathbf{v} c(p, t))=f(p)
\end{equation}
where $c(p,t)$ is the spatio-temporal scalar field describing the concentration of the dispersion; $\lambda$ is the diffusion coefficient; $\mathbf{v}$ is the wave velocity; $f(p)$ is the source input. As mentioned above, given the initial concentration, $c_0$, source input, $f\left( p \right)$, and the wave velocity, $\mathbf{v}$, the pollutant concentration distributed in the spatial area, over time, can be quantified by solving the equation. This procedure can be calculated using common \gls{cfd} solvers. In this work, we adopted the reliable open-source solution from OpenFOAM \autocite{jasak_openfoam_2007}, which is one of the best-known \gls{cfd} solvers. An example of the simulated pollutant dispersion is shown in Fig. \ref{fig:dispersion}.

The physical simulator of the survey \gls{usv} is mainly based on Gazebo. This enables easy integration with a wide range of open-source libraries. In this work, the implementation of a wave-driven \gls{usv} hydrodynamics simulator \autocite{mainwaring_wave_2025}, and ArduPilot, the autopilot software \autocite{ardupilot_community_ardupilot_2025}, are utilised. The hydrodynamics simulator allows for an exceptionally high-fidelity environment. In this simulated environment, the consistency between the hydrodynamics of the \gls{usv} and the pollutant advection is rigorously facilitated by aligning the wave velocity parameters in both the \gls{cfd} solver and the physical simulator. Furthermore, the ArduPilot is used to generate thruster commands based on higher-level navigation waypoints. This allows for the integration of navigation and control systems, enabling the \gls{usv} to autonomously follow waypoints and perform complex manoeuvres.
% In order to incorporate the ArduPilot into the ROS middleware, a dedicated DDS library is used as described in [CITE].
\subsection{ROS-based data interfaces}
In this work, the data interfaces between the simulation module and the \gls{ros}-based source tracking system were strictly constructed to mimic the real-world data collection and passing patterns. The simulated pollutant concentration data were derived from the \gls{cfd} dispersion results and the \gls{usv} position in Gazebo. These data were collected by a `sonde driver' and then passed to the main tracking system through a \gls{ros} interface. This procedure was designed to simulate the behaviour of a multi-parameter sonde \autocite{ma_at_sonde_ros_driver_2025-1, li_ysi_exo_2022}, which is commonly used in marine environments to measure pollutant concentrations. Upon receiving the probed data from the `sonde driver', the main algorithm, introduced in Section \ref{s:ipp}, updates the estimation of the pollution source and calculates the navigation waypoint. This waypoint was sent to the ArduPilot also via a \gls{ros} interface, which eventually drove the thrusters of the \gls{usv}.

% It is also worth noting that the main source tracking algorithm was constructed as a \gls{ros} action server, which provides a \gls{ros} action. A \gls{ros} action, as described by Macenski et al. \cite{macenski_robot_2022}, enables convenient interaction with procedures that require a significant amount of time to return a response. It is widely applied in fields of path planning \cite{macenski_marathon_2020} and manipulation \cite{gorner_moveit_2019}. Constructing the main tracking algorithm as an action server also offers a means of extension for future modular development. Several action servers can be incorporated into a higher-level decision-making logic through behaviour trees \cite{colledanchise_behavior_2018} and state machines \cite{foukarakis_combining_2014}.

\section{Categorical Bayesian source tracking}
\label{s:ipp}

% In this section, the main pollution source tracking system is introduced. 
In the proposed tracking system, an uncertainty-aware estimate of the source location was iteratively updated, based on the probed data passed from the sonde driver. As described in \textcite{vergassola_infotaxis_2007}, the information about the estimation obtained in the next navigation waypoint can be quantified as the expected information gain. The navigation waypoint that maximises the expected information gain was selected and executed by ArduPilot. Throughout the source tracking procedure, the uncertainty levels of the estimates were closely monitored. When the uncertainty levels reached a satisfactory threshold, the server containing the tracking system would respond with the latest estimate as final results.

In line with several works in pollution survey \autocite{liu_bayesian_2024, ma_adaptive_2025, song_novel_2022}, the tracking area was discretised into a set of equally sized grids $g_{i}$. Each grid was attached to a coordinate with respect to a local \gls{neu} frame, $(x_{i}, y_{i})$, and the probability, $p \left( g_{i} \right)$, that the source of the pollution is contained within the grid was recorded. Therefore, the entire grid map was constructed into a discrete probability distribution, which represented the overall estimate of the pollution source location. Instead of \gls{cimb} \autocite{ojeda_information-driven_2021}, the grid map was formulated into a categorical distribution with $k$ categories, where $k$ is the number of grids. This addressed the ambiguity caused by normalising the multivariate Bernoulli and assuming the Bernoulli distributions as collectively independent. Using the categorical distribution, the best estimate was given as the expectation of it:
\begin{align}
    x_s &= \sum_{i} x_i p \left( g_{i} \right) \\
    y_s &= \sum_{i} y_i p \left( g_{i} \right)
\end{align}

On receiving new probed data, a real-time estimate instance was generated using the method as proposed by \textcite{ojeda_information-driven_2021}. In this method, a local estimate was obtained via Gaussian kernels based on the wave direction, $\hat{\textbf{v}} = \frac{\textbf{v}}{\left\| \textbf{v} \right\|_2}$, and the bearing of the latest above-threshold measurement, $\hat{\textbf{r}} = \frac{\textbf{r}}{\left\| \textbf{r} \right\|_2}$:
\begin{align}
    \label{eq:kernel1}
    p_t \left( g_i | z_t = 1 \right) &\sim \mathcal{N} \left( \text{atan2} \left(\hat{\textbf{v}}\left[2\right], \hat{\textbf{v}}\left[1\right] \right), \, \sigma^2_a \right) \\
    p_t \left( g_i | z_t = 0 \right) &\sim \mathcal{N} \left( \text{atan2} \left(\hat{\textbf{r}}\left[2\right], \hat{\textbf{r}}\left[1\right] \right), \, \sigma^2_b \right)
\end{align}
% \eqref{eq:kernel1}
where $z_t$ is the probed data at time $t$; $z_t = 1$ denotes an above-threshold measurement and $z_t = 0$ denotes a below-threshold measurement; $\mathbf{r}$ is a vector pointing from the \gls{usv} to the position of the latest above-threshold measurement; $\sigma^2_a = 0.04s_g$ and $\sigma^2_b = 0.16s_g$ are empirically selected variances, where $s_g$ is the area of a grid. 
Unlike other source-tracking algorithms \autocite{vergassola_infotaxis_2007}, where both the advection speed and direction are required, in \eqref{eq:kernel1}, only the direction of the wave was assumed to be known to the Bayesian inference system, which is easier to satisfy and improves the system robustness.
The local estimate covered only a small neighbourhood of the current grid, so the uncovered grids were assigned with the same probability as the closest grid that was covered by the local estimate. The existing estimate of the source location was then updated based on Bayes' theorem:
\begin{equation}
\label{eq:bayes}
    p \left( g_i | z_{1:t} \right) \varpropto p \left( g_i | z_{t} \right)p \left( g_i | z_{1:t-1} \right)
\end{equation}

% \begin{figure*}[htbp]
% \centerline{\includegraphics[width=0.9\textwidth]{figs/prob_loc.pdf}}
% \caption{Source location probability map at different time steps}
% \label{fig:prob_loc}
% \end{figure*}

% After the updated estimate of the pollution source location was obtained, a new navigation waypoint was selected to guide the \gls{usv} cruising towards the next search area:
Based on the latest posterior, the next navigation waypoint was selected as:
\begin{align}
    g^* = \arg \max_{g \in \varPsi} IG\left( g | p_{t-1} \right)
\end{align}
where $\varPsi$ is the set of candidate waypoints. These waypoints were grid centres of a defined neighbourhood encasing the current \gls{usv} position. 
% In this work, the workspace consisted of 100 $\times$ 50 grids, and the framework was executed on a standard laptop with an Intel$^\circledR$ Core$^{\text{TM}}$ Ultra 7 Processor 155H and 32 GB of RAM. The size of the neighbourhood was empirically selected as 11 grid $\times$ 11 grid to balance the computational burden and the search horizon. 
The information gain, $IG\left( g | p_{t-1} \right)$, was formulated as the \gls{kld} between the future and the existing estimate. Although the use of \gls{kld} as information gain was partially based on the existing works \autocite{ojeda_information-driven_2021, lopes_information_2022}, the \gls{cimb}-based probabilistic modelling was substituted with a categorical distribution as mentioned previously, to better represent the discrete Bayesian source localisation. The information gain derived from the categorical-distribution estimate was given as:
\begin{align}
    IG\left( g | p_{t-1} \right)& = p_{t-1}\left(g\right) \sum_{i=1}^k p_{t,a}\left( g_i \right) \ln \frac{p_{t,a}\left( g_i \right)}{p_{t-1}\left( g_i \right)} \nonumber \\
    & + \left(1 - p_{t-1}\left(g\right)\right) \sum_{i=1}^k p_{t,b}\left( g_i \right) \ln \frac{p_{t,b}\left( g_i \right)}{p_{t-1}\left( g_i \right)}
\end{align}
where $p_{t,a}$ and $p_{t,b}$ are the future estimates of the source location given the above-threshold and below-threshold measurements, respectively. These estimates can be derived by updating the latest belief with the corresponding assumption of the probed data using \eqref{eq:bayes}:
\begin{align}
    \ln p_{t,z}\left( g_i \right) = \ln \frac{p_{t-1}\left( g_i \right) p_t \left( g_i | z_t \right)}{\sum_{j=1}^k p_t \left( g_j | z_t \right)} - \ln Z
\end{align}
where 
% $Z=\sum_i^k \frac{p_{t-1}\left( g_i \right) p_t \left( g_i | z_t \right)}{\sum_{j=1}^k p_t \left( g_j | z_t \right)}$.
\begin{align}
    Z=\sum_i^k \frac{p_{t-1}\left( g_i \right) p_t \left( g_i | z_t \right)}{\sum_{j=1}^k p_t \left( g_j | z_t \right)}
\end{align}

In order to closely monitor the uncertainty level, a novel step was to use credible intervals of the source location to quantify the uncertainty level during the tracking. The credible interval $\left[q_l, q_u \right]$, is a Bayesian statistics concept \autocite{eberly_estimating_2003}, which is defined as that the source location has a probability $\gamma$ to be within it:
\begin{equation}
    \sum_{i = q_l} ^{q_u} p \left( g_i | z_{1:t} \right) \geq \gamma
\label{eq:ci}
\end{equation}

The $\gamma$-\gls{sci} was selected from various credible interval variants. The $\gamma$-\gls{sci} $\left[q_l^{\prime}, q_u^{\prime} \right]$ is defined as:
\begin{align}
    \forall \left[q_l, q_u \right] &\in \left\{ \left[q_l, q_u \right] \mid \sum_{i = q_l} ^{q_u} p \left( g_i | z_{1:t} \right) \geq \gamma \right\} \nonumber \\
    \left[q_l^{\prime}, q_u^{\prime} \right] &= \arg \min_{\left[q_l, q_u \right]} \left( q_u - q_l \right) 
\end{align}
which facilitates a precise representation of the uncertainty level by focusing on the area with the highest probability mass. Besides providing a real-time representation of uncertainty levels, the \gls{sci} was also used to determine the termination condition of the tracking system. When $q_u^{\prime} - q_l^{\prime} \leq \tau$, where $\tau$ is the \gls{sci} threshold, the action server will respond to the request client with the latest best estimate of the pollution source location. The threshold for a 99\%-\gls{sci} was given as the length of two grids in this work. The $\gamma = 99\%$ and the extremely small threshold value were selected based on the fact that the simulated environment is inevitably more ideal compared with reality, though high-fidelity simulation has been used. When this is applied to real \glspl{usv}, we recommend suitable adjustments of these parameters.

\section{Results}
\begin{table*}[htbp]
\caption{Configuration of the validation scenarios}
\label{tab:config}
\centering
\begin{tabular}{@{}lccc@{}}
\toprule
Scenario                  & a              & b    & c          \\ \midrule
Source location (m)       & (2.5, 2.5)     & (102.5, -52.5) & (-102.5, -52.5) \\
Source release rate (kg/s)
    & 2.5            & 2.5      & 2.5   \\
Wave velocity $\mathbf{v}$ (m/s)       & $\left[1.2247, 1.2247\right]^\intercal$ & $\left[-1.2247, 1.2247\right]^\intercal$  &$\left[1.2247, 1.2247\right]^\intercal$\\
USV starting position 1 (m) & (60, 60)     & (0, 50)    & (0, 50)\\
USV starting position 2 (m) & (120, 120)     & (-120, 120)    & (120, 120)\\
\bottomrule
\end{tabular}
\end{table*}

The proposed framework aims to enable uncertainty-aware autonomous pollution source tracking using a survey \gls{usv} within the high-fidelity marine environment described in \ref{ss:simulator}. This section presents validation results of the proposed tracking system in three simulated pollution dispersion scenarios. For each scenario, two different \gls{usv} starting points were tested.

\subsection{Configuration}
All dispersion scenarios used a 500 m $\times$ 250 m workspace. As mentioned in \ref{s:ipp}, the workspace was discretised into 100 $\times$ 50 equally spaced grids, each measuring 5 m $\times$ 5 m. In the simulation, the origin of the local \gls{neu} frame was positioned at the centre of the workspace. The wave field was assumed to be uniform within the workspace, which is a common approach in practical wave velocity measurements \autocite{cao_behavior_2022, hutchinson_source_2019}, parameterised with $\mathbf{v} \in \mathbb{R}^{2}$. Fluorescein, an organic dye, was selected as the simulated pollutant with a diffusion coefficient of $\lambda = 4.9\times 10^{-10} \, \text{m}^2\text{/s}$ \autocite{rani_rapid_2005}. The three scenarios were devised with different source locations, wave velocities and starting positions of the survey \gls{usv}. Configuration details are provided in Tab. \ref{tab:config}. Each scenario was configured with two different \gls{usv} starting positions. Compared to the first positions, the second positions were located farther away from the source and were therefore more challenging.

The proposed framework was executed on a standard laptop with an Intel$^\circledR$ Core$^{\text{TM}}$ Ultra 7 Processor 155H and 32 GB of RAM. The size of the neighbourhood $\varPsi$ was empirically selected as 11 grid $\times$ 11 grid to balance the computational burden and the search horizon. The termination of the tracking was controlled by the $\gamma$-\gls{sci} of the estimated distribution. In these validations, the $\gamma$-\gls{sci} threshold was $\tau = 10 \text{ m}$. The validation trials that cannot satisfy the $\gamma$-\gls{sci} threshold within 600 seconds were forcibly terminated and considered failed.

\subsection{Uncertainty-aware source tracking}

\begin{figure*}[htbp]
    \renewcommand{\thesubfigure}{a-\roman{subfigure}}
\centering
    \begin{subfigure}[t]{0.49\textwidth}
        \centering
        \includegraphics[width=\textwidth]{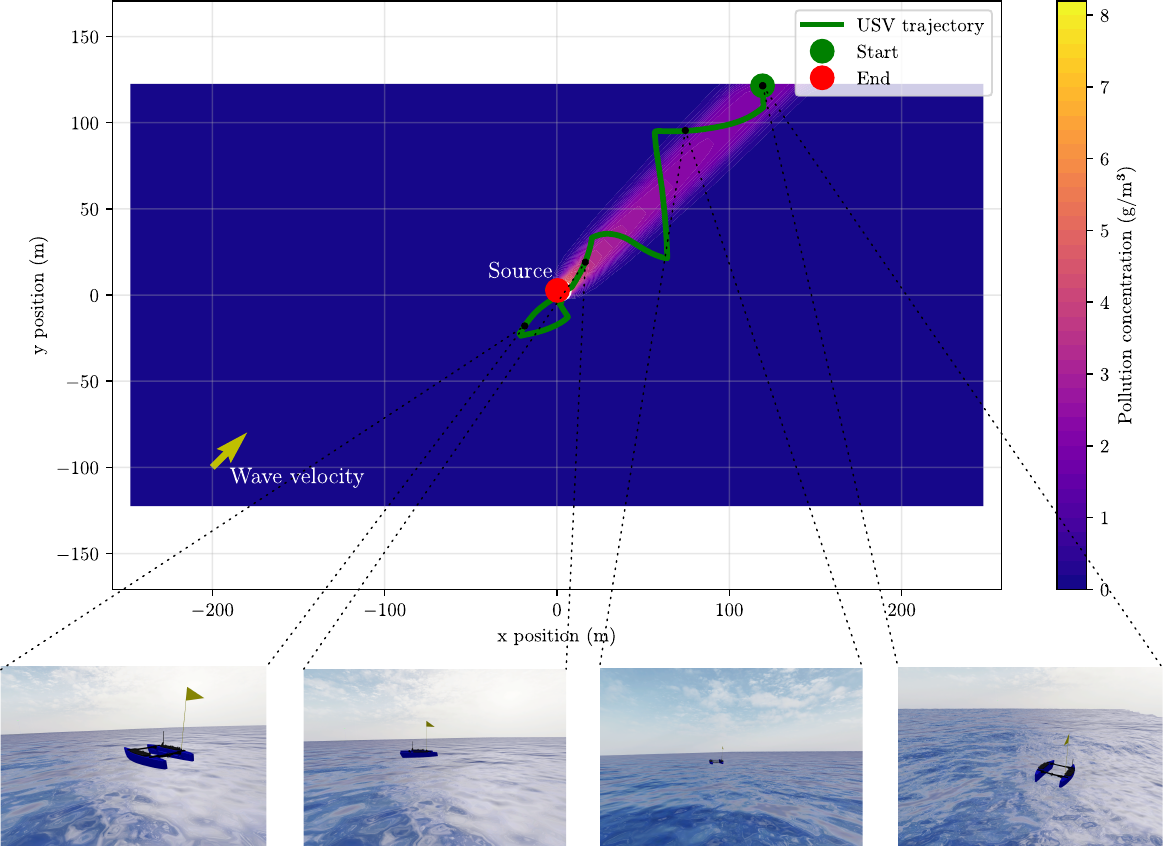}
        \caption{Active tracking trajectory in simulated environment}
        \label{subfig:s2_st1_tr}
    \end{subfigure}
    \begin{subfigure}[t]{0.49\textwidth}
        \centering
        \includegraphics[width=\textwidth]{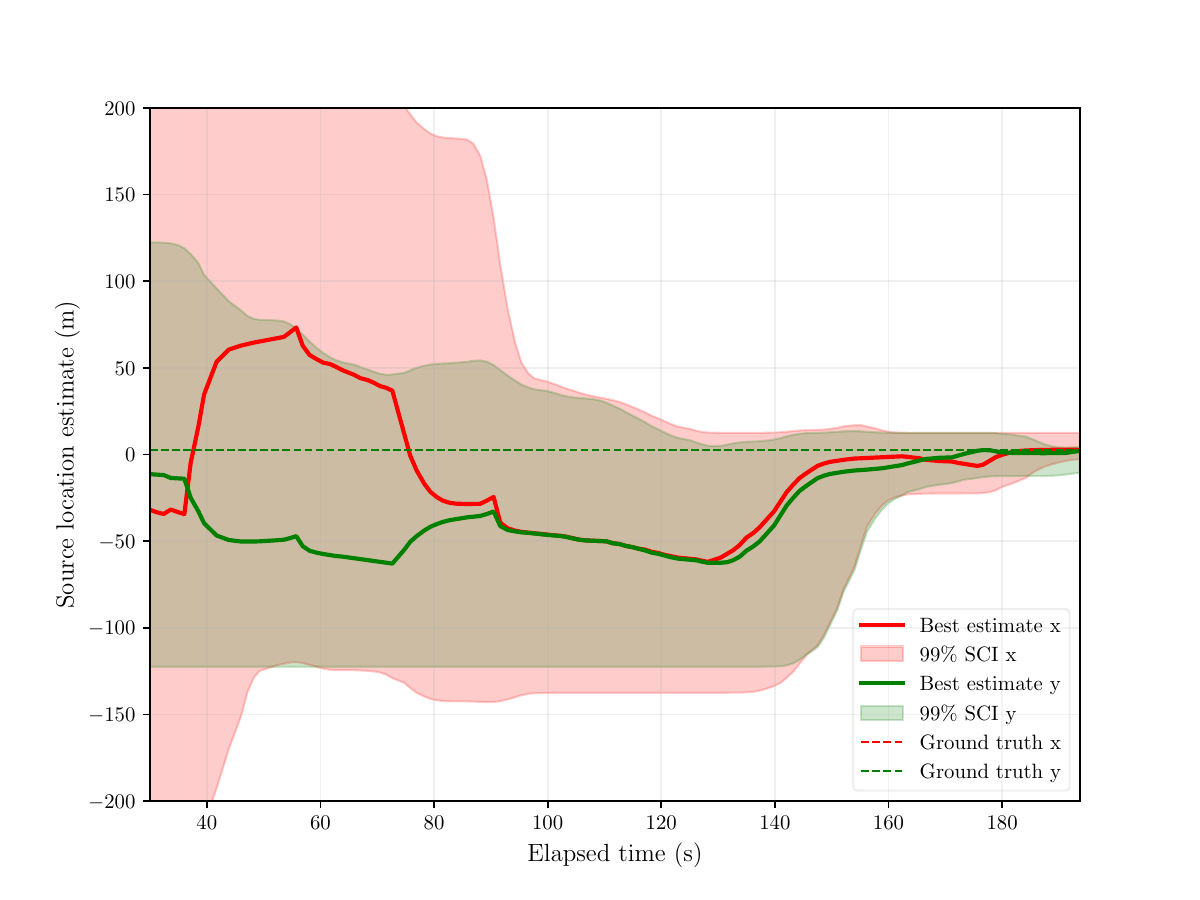}
        \caption{Source location estimate and SCI}
        \label{subfig:s2_st1_ci}
    \end{subfigure}

\vspace{1em}
\setcounter{subfigure}{0} % Reset subfigure counter to start from 'a' again
\renewcommand{\thesubfigure}{b-\roman{subfigure}}
\centering
    \begin{subfigure}[t]{0.49\textwidth}
        \centering
        \includegraphics[width=\textwidth]{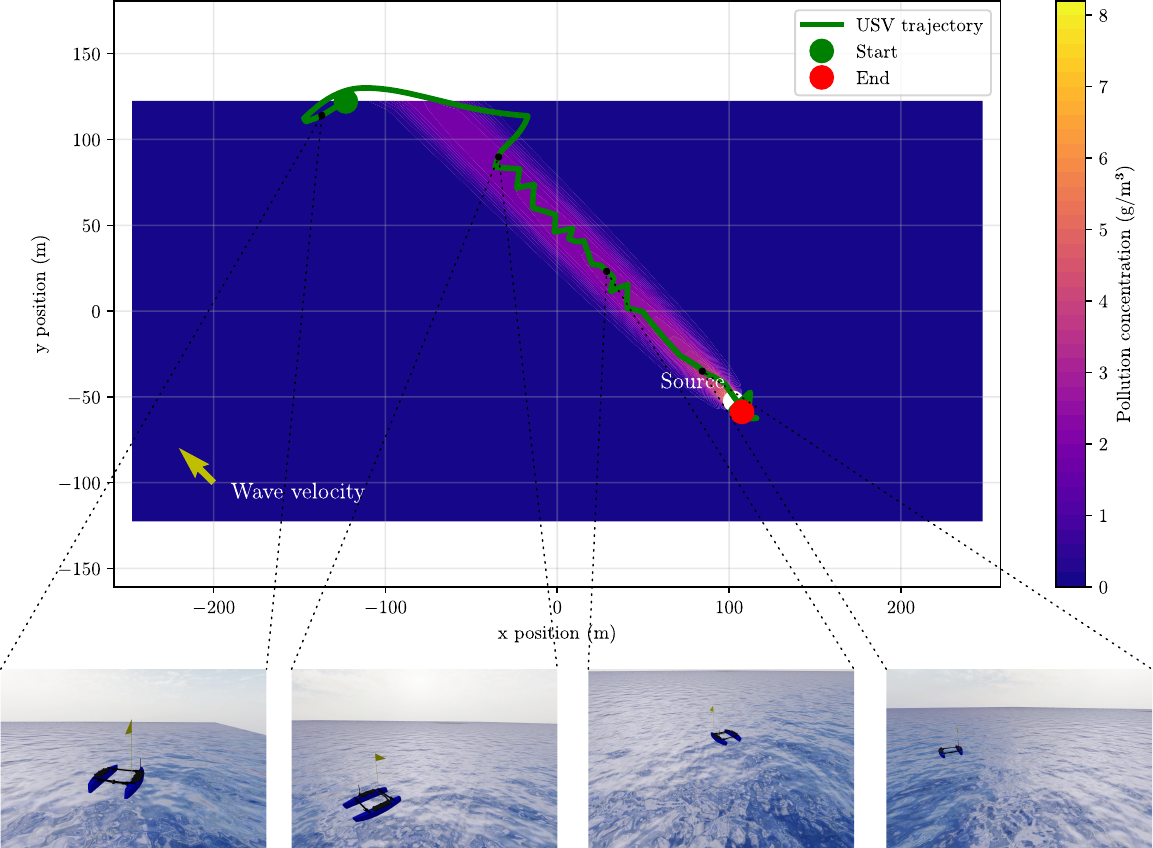}
        \caption{Active tracking trajectory in simulated environment}
        \label{subfig:s3_st1_tr}
    \end{subfigure}
    \begin{subfigure}[t]{0.49\textwidth}
        \centering
        \includegraphics[width=\textwidth]{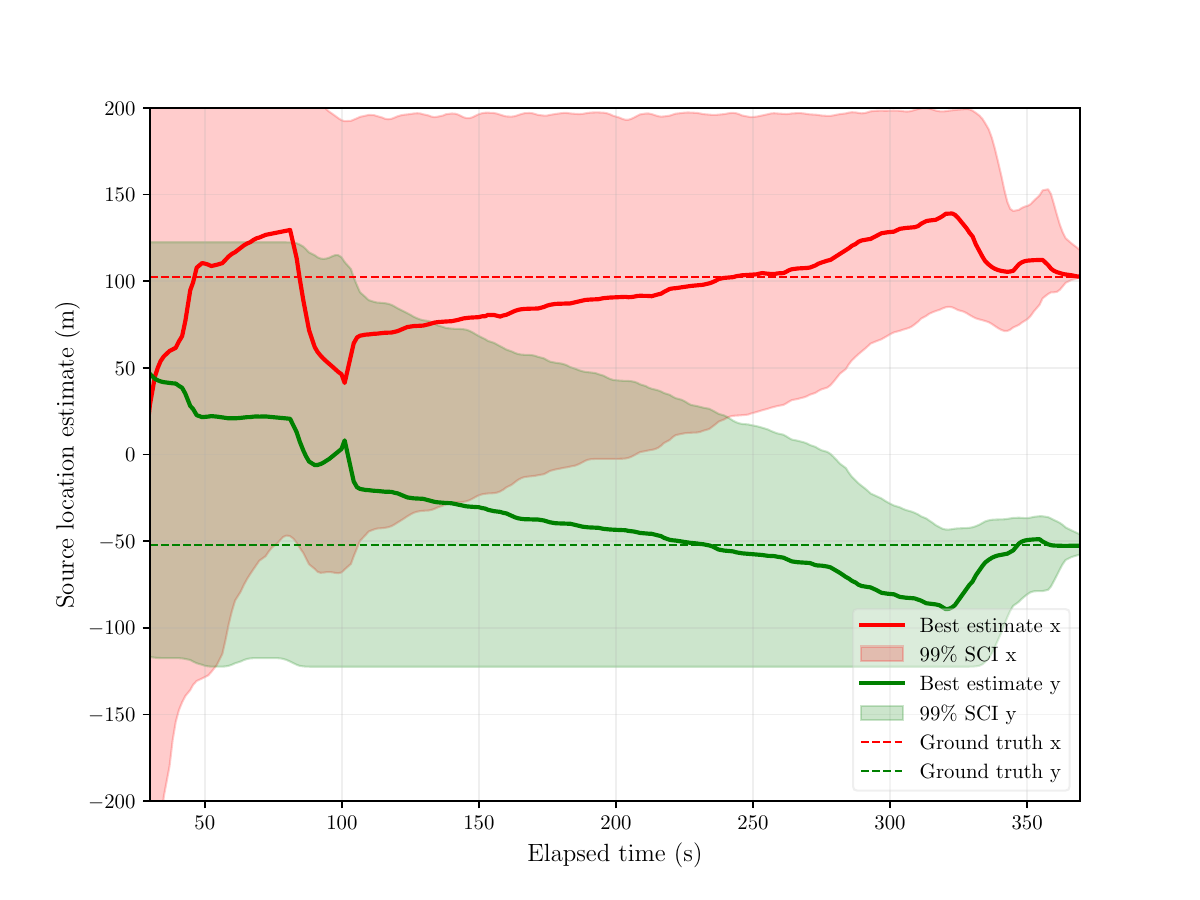}
        \caption{Source location estimate and SCI}
        \label{subfig:s3_st1_ci}
    \end{subfigure}
\caption{(a) Scenario a-2 of the validation. (b) Scenario b-2 of the validation. (i) The trajectories of the \gls{usv} during the tracking procedure. The dispersion plume is time-varying, and the plume shown in the figure is the one at the end of the tracking. (ii) The evolution of the source location estimate and the $\gamma$-\gls{sci} during the tracking procedure. The shaded area represents the estimation uncertainty, which decreases as the tracking proceeds.}
\label{fig:trials}
\end{figure*}

In all scenario variants, the proposed source tracking system successfully identified the pollution source without human intervention in most trials. We take two trial instances from a-2 and b-2 as examples. Their detailed data, along with their trajectories and uncertainty levels, are provided in Figs. \ref{fig:trials}.

At the beginning, the survey \gls{usv} cruised along the pollutant plume. On the whole, it moved to the upwave direction, as shown in Fig. \ref{subfig:s2_st1_tr} and \ref{subfig:s3_st1_tr}. Towards the end, the survey \gls{usv} overshot the pollution source, reaching an area with little pollutant. At this stage, the tracking system guided the \gls{usv} to conduct a detailed search in the high probability mass area.
% ; an example of a series of source location probability maps is shown in Fig. \ref{fig:prob_loc}. 
This detailed search typically included numerous re-entering and leaving the pollutant plume. The $\gamma$-\gls{sci} noticeably dropped during this search, as shown in Fig. \ref{subfig:s2_st1_ci} and \ref{subfig:s3_st1_ci}, and eventually reached the 10 m threshold. In practical applications, by monitoring the $\gamma$-\gls{sci} in real-time, operators can have a clear understanding of the uncertainty level of the current estimate.

In order to better mimic the real-world environment, the random seeds used by the simulation models were not fixed, which introduced stochasticity into the simulated model. The exact time taken for the \gls{usv} firmware to arm the motors and prepare to navigate can also range from 5 to 10 seconds. During this time interval, the \gls{usv} drifts from the configured starting position because of the wave, which leads to a small amount of randomisation of the real starting position of the tracking algorithm. Therefore, in each scenario variant, a set of trials was conducted to rigorously examine the average performance. More details about the sampling and trials can be found in the next section.

% In scenario (\textbf{b})-2, the time consumed by the tracking system was overall longer than in scenario (\textbf{a})-2. Although the Euclidean distance between the source and the \gls{usv} starting point was slightly longer in (\textbf{b})-2, the difference between the consumed times could be mainly attributed to the starting position of the survey \gls{usv}; in scenario (\textbf{b})-2, the starting position was located outside the pollutant plume. It was recognised that, prior to tracking within the plume, the \gls{usv} conducted a considerable amount of exploration to obtain the first above-threshold pollution concentration. This was unlike scenario (\textbf{a})-2, where the starting position was within the pollutant plume.

% It should, however, be noted that if the \gls{usv} starting position was exceptionally far away from the pollutant plume, there would be a good probability that the tracking system failed to identify the source, for example, when the starting position was placed on the upwave direction with respect to the source. This was due to the tracking algorithm relying on above-threshold pollution concentration measurements to correct itself when the \gls{usv} left the plume. The exploration procedure that discovered the initial above-threshold measurement was out of the scope of this article. Nevertheless, as suggested in \ref{s:framework}, an exploration action server can be easily integrated with the proposed framework using the behaviour tree in future work.

\subsection{Comparative study}
\begin{table*}[htbp]
  \centering
  \caption{Trial success rates across experimental scenarios}
  \label{tab:success_rate}
  \begin{small}
  \begin{tabular}{lccccccc}
    \toprule
    \multirow{2}{*}{Method} & \multicolumn{2}{c}{Scenario a} & \multicolumn{2}{c}{Scenario b} & \multicolumn{2}{c}{Scenario c} & \multirow{2}{*}{Overall} \\
    \cmidrule(lr){2-3} \cmidrule(lr){4-5} \cmidrule(lr){6-7}
    & a-1 & a-2 & b-1 & b-2 & c-1 & c-2 & \\
    \midrule
    \textbf{Proposed} & 100.0\% & 100.0\% & 100.0\% & 75.0\% & 100.0\% & 100.0\% & \textbf{95.8\%} \\
    CIMB              & 100.0\% & 75.0\%  & 100.0\% & 75.0\% & 100.0\% & 100.0\% & 91.7\% \\
    Bessel            & 60.0\%  & 100.0\% & 100.0\% & 50.0\% & 100.0\% & 0.0\%   & 68.3\% \\
    \bottomrule
  \end{tabular}
  \end{small}
\end{table*}
\begin{figure}[htbp]
    \centering
    \includegraphics[width=\linewidth]{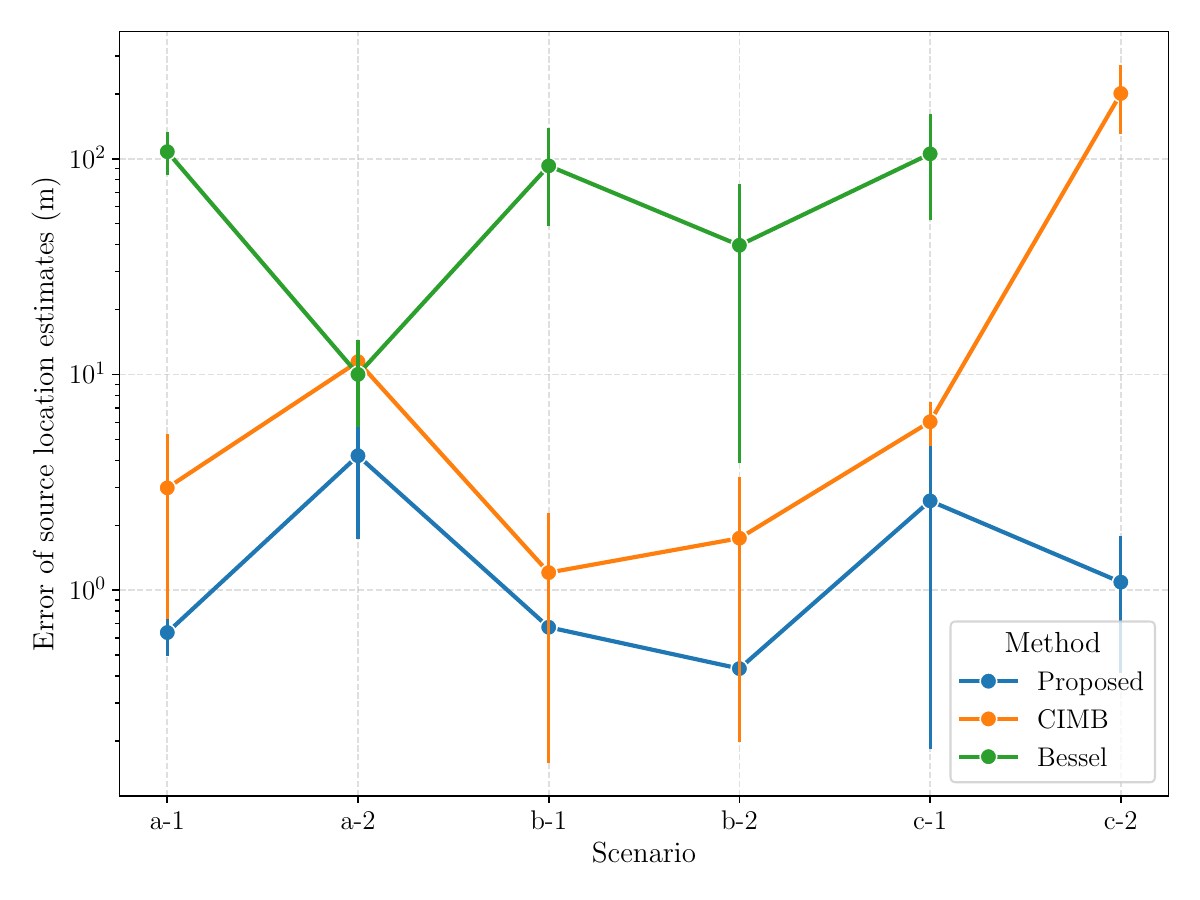}
    \caption{Method performance comparison in terms of source location error across different scenarios. The data points represent the mean error of successful trials, and the error bars represent the standard error of the mean. The plot is in logarithmic scale for better visualisation.}
    \label{fig:int_method_comparison}
\end{figure}
\begin{figure}[htbp]
    \centering
    \includegraphics[width=\linewidth]{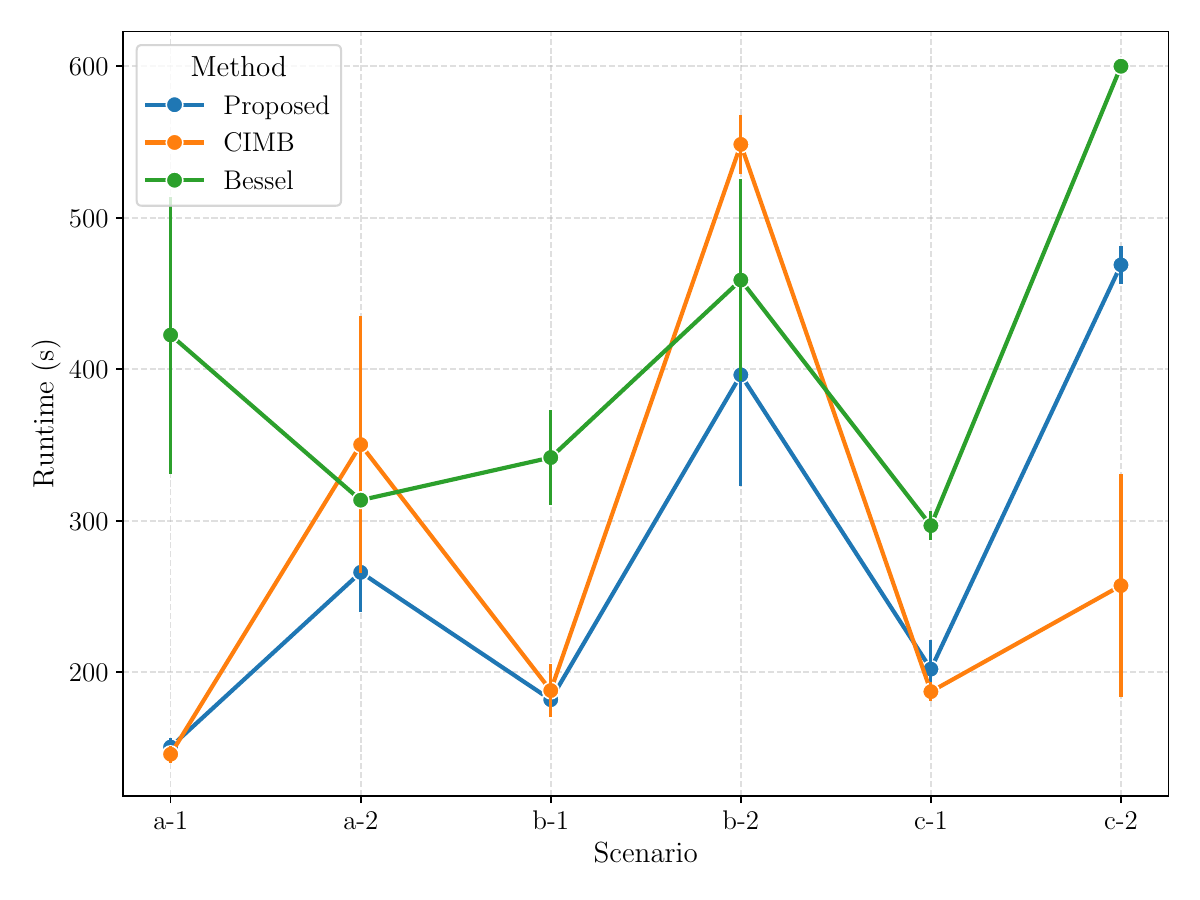}
    \caption{Runtime comparison of the methods across different scenarios. The data points represent the mean runtime and the error bars represent the standard error. A 600 s runtime indicates a failed trial.}
    \label{fig:int_method_runtime}
\end{figure}
\begin{table}[htbp]
  \centering
  \caption{Two-way ANOVA results for estimation accuracy comparison between methods and scenarios}
  \label{tab:anova_results}
  \begin{small}
  \begin{tabular}{lcc}
    \toprule
      & $p$ & $\eta_p^2$ \\
    \midrule
    \textbf{Proposed vs. CIMB} & & \\
    Method (M) & 0.0038 & 0.227 \\
    Scenario (S) & $<$0.001 & 0.531 \\
    Interaction (M $\times$ S) & $<$0.001 & 0.535 \\
    \midrule
    \textbf{Proposed vs. Bessel} & & \\
    Method (M) & $<$0.001 & 0.408 \\
    Scenario (S) & $<$0.001 & 0.606 \\
    Interaction (M $\times$ S) & 0.233 & 0.195 \\
    \bottomrule
  \end{tabular}
  \end{small}
\end{table}

To examine the proposed probability modelling, we compared pollution source location estimation error and tracking runtime between the proposed categorical distribution and the \gls{cimb} baseline \autocite{ojeda_information-driven_2021}. The result using the classic Bessel plume infotaxis \autocite{vergassola_infotaxis_2007} is also reported in this section, serving as a general baseline. This comparison was conducted in all six scenario variants, as described in Tab. \ref{tab:config}.

Due to the fact that the original work of the Bessel plume infotaxis \autocite{vergassola_infotaxis_2007} lacks an explicit source declaration or search termination mechanism, in this comparative study, we focused on the influence of the different Bayesian update strategies and, therefore, the tracking termination for all methods was determined by the proposed SCI under same configuration.

To ensure statistical power while accounting for potential algorithmic failures in complex scenarios, a dynamic sampling strategy was employed. Each method-scenario combination was initially tested with four independent trials. If the number of successful trials was less than three, additional trials were conducted until a minimum of three valid data points was reached, or a maximum 10-attempt limit was exhausted, allowing for robust variance estimation.

Before analysing the quantitative performance metrics, the robustness of each method was assessed. As summarised in the Tab. \ref{tab:success_rate}, the proposed method demonstrated exceptional environmental adaptability, achieving an overall success rate of 95.8\%, with 100\% success in five out of six scenarios. The CIMB method also showed stable robustness with a 91.7\% overall success rate. In contrast, the Bessel method exhibited severe vulnerability to environmental complexity. Its overall success rate dropped to 68.3\%, and it completely failed to produce any valid results in the most challenging scenario, c-2.

For the successful trials, the error data of the source location estimates were analysed using an unbalanced two-way \gls{anova}. As detailed in Tab. \ref{tab:anova_results}, the results revealed a highly significant main effect for both Method and Scenario ($p < 0.05$). The Method effect had a large effect size, indicating that the choice of method significantly influenced estimation accuracy. The Scenario effect was even more pronounced, reflecting the varying levels of difficulty across different simulation scenarios. To further untangle the analysis, the interaction plot (Fig. \ref{fig:int_method_comparison}) was provided alongside the two-way \glspl{anova} conducted for method pairs.

Prior to the \glspl{anova}, Levene's tests validated the homogeneity of variances across groups ($p > 0.05$). The \gls{anova} results revealed that the proposed method exhibits a statistically significant main effect over Bessel ($p < 0.001, \eta_p^2 = 0.408$). Furthermore, the interaction term is not significant ($p = 0.233$). As is visually clear in Fig. \ref{fig:int_method_comparison}, the performance curve of the proposed method remains strictly below that of the Bessel method across all scenarios. This provides compelling evidence that the proposed method universally and robustly outperforms Bessel, regardless of scenario variations. While the proposed method maintains a significant overall performance advantage over CIMB ($p = 0.0038, \eta_p^2 = 0.227$), a significant interaction effect is present ($p < 0.001, \eta_p^2 = 0.535$). The Interaction Plot clearly illustrates the nature of this interaction: in simpler environments (e.g., b-1), the performance lines of the proposed method and CIMB are to some extent aligned. However, as the scenario complexity increases (particularly in c-2), the CIMB curve suffers from severe degradation, whereas the proposed method's curve remains consistent. The overall variance of the sampled results from the proposed method is also less than the CIMB, indicating its robustness. Referring to Fig. \ref{fig:int_method_runtime}, though, in some complicated scenarios, the proposed method requires a longer search runtime for higher accuracy; however, the overall runtime level is still in line with the trend of other baseline methods.

\section{Discussion and conclusion}
Prior work has demonstrated the effectiveness of active tracking in pollution monitoring \autocite{ma_adaptive_2025}. Since the initial concept of infotaxis \autocite{vergassola_infotaxis_2007} was proposed to guide navigation without relying on the gradient, there have been a number of research works further pushing the boundaries. For example, \textcite{ojeda_information-driven_2021} presented a method for indoor gas source localisation using Bayesian estimation and \gls{ipp}, while \textcite{bayat_optimal_2016} reported a particle-filter-based active tracking system for marine pollution. However, a significant gap remains: few have investigated the effect of the underlying probability model of the Bayesian inference. Additionally, there is a need for a specialised simulation pipeline for marine pollution monitoring.
% in the field of \gls{ipp}, few studies have constructed a dedicated high-fidelity simulation pipeline.

To address this, this paper proposed an uncertainty-aware active marine pollution source tracking framework based on categorical Bayesian estimation. 
% We validated the proposed system in a high-fidelity, \gls{ros}-compatible simulation environment across various scenarios, 
% The proposed framework demonstrating its reliable and superior performance compared to other baseline methods. 
An extensive analysis of the influence of probability modelling in Bayesian inference revealed the superiority of categorical distribution over other baseline methods in terms of both reliability and performance. 
This work extends the findings of \textcite{bayat_optimal_2016} by developing a more comprehensive Bayesian-based active tracking system and validating it in a higher-fidelity environment. 
To our knowledge, this is the first framework to achieve active pollution source tracking validated in a high-fidelity marine environment. The robust performance demonstrated in these validations supports its potential for future hardware implementation.

This work has several limitations that point to future research directions. Currently, the framework is limited to the source tracking function; incorporating an exploration module would be beneficial to ensure initial, above-threshold measurements can be reliably obtained. We also acknowledged the limitation of the current \gls{cfd} model, which can not fully represent the extremely complicated mixing and turbulent factors. Future work should incorporate more sophisticated \gls{cfd} models and real-world dispersion experimental data. Finally, while the simulator was constructed with high fidelity to serve as a strong proof of concept, future work will involve implementing the framework in real-world hardware to validate its performance on a physical system.

%% For bibtex:
% \bibliographystyle{IEEEtran}
% \bibliography{IEEEabrv,references}

% If switch to biblatex
\renewcommand*{\UrlFont}{\rmfamily}
\renewcommand*{\bibfont}{\footnotesize}
% \setquotestyle{american}
\printbibliography

\end{document}